\documentclass{ecai} 
\usepackage{latexsym}
\usepackage{amssymb}
\usepackage{amsmath}
\usepackage{amsthm}
\usepackage{booktabs}
\usepackage{enumitem}
\usepackage{graphicx}
\usepackage{xcolor}

\makeatletter
\let\MYcaption\@makecaption
\makeatother

\usepackage{subfigure}
\makeatletter
\let\@makecaption\MYcaption
\makeatother

\usepackage{multirow}

\newcommand{\block}[1]{\smallskip\noindent{\textbf{#1}}}

\theoremstyle{definition}
\newtheorem{definition}{Definition} %[section]
\newtheorem{theorem}{Theorem} %[section]
\newtheorem{lemma}[theorem]{Lemma}
\newcommand*{\QED}{\hfill\ensuremath{\blacksquare}}

\usepackage{bm}

\DeclareMathOperator{\sign}{sgn}

\newcommand{\argmin}{\mathop{\rm arg~min}\limits}

\newcommand*{\enorm}[1]{\left\| #1 \right\|_2}

\newcommand*{\nvector}[1]{\bm{{#1}}}

\newcommand*{\nset}[1]{\mathcal{\uppercase{#1}}}

\newcommand{\bx}{\nvector{x}}

\newcommand\eqlab[1]{\label{eq:#1}}
\renewcommand*{\eqref}[1]{Eq. (\ref{eq:#1})}

\newcommand\seclab[1]{\label{sec:#1}}
\newcommand\secref[1]{Section \ref{sec:#1}}

\newcommand\applab[1]{\label{app:#1}}
\newcommand\appref[1]{Appendix \ref{app:#1}}

\newcommand\figref[1]{{Figure \ref{fig:#1}}}
\newcommand\tabref[1]{{Table \ref{tab:#1}}}

\newcommand*{\deflab}[1]{\label{def:#1}}
\newcommand*{\defref}[1]{{Definition \ref{def:#1}}}

\newcommand\lemlab[1]{\label{lem:#1}}
\newcommand\lemref[1]{{Lemma \ref{lem:#1}}}

\DeclareMathOperator{\doubleN}{\mathbb{N}}
\DeclareMathOperator{\doubleE}{\mathbb{E}}

\DeclareMathOperator{\doubleI}{\mathbb{I}}
\DeclareMathOperator{\doubleR}{\mathbb{R}}

\DeclareMathOperator{\scriptH}{\mathcal{H}}

\DeclareMathOperator{\scriptX}{\mathcal{X}}
\DeclareMathOperator{\scriptY}{\mathcal{Y}}

\newcommand{\be}{\nvector{e}}
\newcommand{\ba}{\nvector{a}}

\DeclareMathOperator{\agree}{Agree}

\DeclareMathOperator{\btc}{BTC}
\DeclareMathOperator{\bcex}{BCX}
\DeclareMathOperator{\bcx}{BCX}

\DeclareMathOperator{\argsort}{Argsort}
\DeclareMathOperator{\sort}{Sort}

\DeclareMathOperator{\abso}{Abs}
\DeclareMathOperator{\rank}{rank}
\DeclareMathOperator{\topfeat}{TopFeat}

\DeclareMathOperator{\featagree}{FtrAgr}
\DeclareMathOperator{\rankagree}{RnkAgr}
\DeclareMathOperator{\signagree}{SgnAgr}
\DeclareMathOperator{\signedrankagree}{SgnRnkAgr}
\DeclareMathOperator{\normdisagree}{NormDisagree}

\newcommand{\lfeat}{\ell_{\mathrm{Ftr}}}
\newcommand{\lrank}{\ell_{\mathrm{Rnk}}}
\newcommand{\lsign}{\ell_{\mathrm{Sgn}}}
\newcommand{\lsignedrank}{\ell_{\mathrm{SgnRnk}}}

\newcommand{\Lfeat}{L_{\mathrm{Ftr}}}

\begin{document}

%%%%%%%%%%%%%%%%%%%%%%%%%%%%%%%%%%%%%%%%%%%%%%%%%%%%%%%%%%%%%%%%%%%%%%%%

\begin{frontmatter}

%%% Use this command to specify your submission number.
%%% In doubleblind mode, it will be printed on the first page.

\paperid{1678} 

%%% Use this command to specify the title of your paper.

\title{Backward Compatibility in Attributive Explanation and Enhanced Model Training Method}

%%% Use this combinations of commands to specify all authors of your 
%%% paper. Use \fnms{} and \snm{} to indicate everyone's first names 
%%% and surname. This will help the publisher with indexing the 
%%% proceedings. Please use a reasonable approximation in case your 
%%% name does not neatly split into "first names" and "surname".
%%% Specifying your ORCID digital identifier is optional. 
%%% Use the \thanks{} command to indicate one or more corresponding 
%%% authors and their email address(es). If so desired, you can specify
%%% author contributions using the \footnote{} command.

\author[A]{\fnms{Ryuta}~\snm{Matsuno}\thanks{Email: ryuta-matsuno@nec.com.}}

% \author{\fnms{Second}~\snm{Author}}
% \author{\fnms{Third}~\snm{Author}} 

\address[A]{NEC Corporation}

%%% Use this environment to include an abstract of your paper.
% \onecolumn
\begin{abstract}
Model update is a crucial process in the operation of ML/AI systems.
While updating a model generally enhances the average prediction performance, it also significantly impacts the explanations of predictions.
In real-world applications, even minor changes in explanations can have detrimental consequences.
To tackle this issue, this paper introduces BCX, a quantitative metric that evaluates the backward compatibility of feature attribution explanations between pre- and post-update models.
BCX utilizes practical agreement metrics to calculate the average agreement between the explanations of pre- and post-update models, specifically among samples on which both models accurately predict.
In addition, we propose BCXR, a BCX-aware model training method by designing surrogate losses which theoretically lower bounds agreement scores.
Furthermore, we present a universal variant of BCXR that improves all agreement metrics, utilizing L2 distance among the explanations of the models.
To validate our approach, we conducted experiments on eight real-world datasets, demonstrating that BCXR achieves superior trade-offs between predictive performances and BCX scores, showcasing the effectiveness of our BCXR methods.
\end{abstract}

\end{frontmatter}

%%%%%%%%%%%%%%%%%%%%%%%%%%%%%%%%%%%%%%%%%%%%%%%%%%%%%%%%%%%%%%%%%%%%%%%%

\section{Introduction}
For effective operation of machine learning (ML) systems (i.e., MLOps), model updates are essential  to exploit newly collected data and to adopt the changes in data~\cite{mlops2022kreuzberger,Ruf2021DemystifyingMA,mlops2022testi,mlops2022symeonidis}.
Model updates basically replace an old model (i.e., a pre-update model) with a new model (i.e., a post-update model) trained using more recent and/or larger amounts of data.
Typically, this leads to an improvement in the average prediction performance, but local prediction performance may worsen.
Backward compatibility metrics have been proposed to assess these performance degradation~ \cite{btc2019bansai,nfr2021yan,bec2020srivastava,gbc2022sakai, abcd2023matsuno}.
Furthermore, backward-compatibility-aware retraining methods for model updates have been developed~\cite{btc2019bansai, nfr2021yan, gbc2022sakai, abcd2023matsuno}, revealing that there is a trade-off between backward compatibility and prediction performance of a new model over the old model.

While predictive performance is important for ML models, there are other important demands as well; explainability is one of them, which is often as crucial as predictive performance for sensitive and critical domains, such as healthcare and security.
Recently, explanation methods for ML models, a.k.a. XAI (eXplainable AI), have been actively researched, and various post-hoc and model-agnostic attributive explanation methods~\cite{fae2024wang} have been proposed, including LIME~\cite{lime2016}, Anchors~\cite{anchors2018}, and SHAP~\cite{shap2017,treeshap2018}.
% These explanation methods explain a prediction based on the relevance of each feature of the corresponding input.

Ensuring that the explanations of the new model align with ones of the old model is crucial in real-world applications.
Even though the average prediction performance is improved, the practitioners might hesitate to adopt a new model if it presents different explanations, as this can lead to confusion regarding the real use of the prediction with explanation.
Typically, users perceive the new model as less reliable when they are already familiar with the behavior of the old model~\cite{btc2019bansai}.
While a few studies have examined the disagreement between different explanation methods~\cite{order2021, disagree2022krishna, flora2022comparing1, flora2022comparing2},  the compatibility of explanations during model updates has yet to be explored.

In this study, we introduce a new metric called BCX (Backward Compatibility of eXplanation) to assess the consistency of attributive explanations between old and new models using four practical top-$k$ feature-based agreement metrics~\cite{disagree2022krishna}.
BCX calculates the average agreement of explanations between the old and new models for samples where both models make correct predictions, providing a measure of compatibility in explanations alongside predictive performance.
We then propose BCXR (BCX-aware Retraining) methods.
%, which utilize each of the four agreement metrics.
Since the agreement metrics used in BCX themselves are not differentiable, we propose differential surrogate losses that have theoretical validity for substitution.
Additionally, we present a universal variant of BCXR that can improve the compatibility regardless of the choice of agreement metrics.
To evaluate the effectiveness of our methods, we conduct experiments on eight real-world datasets.
The results demonstrate that BCXR achieves a better trade-offs between BCX scores and predictive performance, thus showing promising efficacy.
Notably, we observe that when the number of features is large, BCXR even outperforms retraining without considering BCX in terms of the predictive performance.
Overall, this study provides a method to evaluate and enhance the compatibility of explanations during model updates, contributing to the establishment of trustworthy and responsible MLOps.

\setlist[enumerate]{leftmargin=2.5em}
To summarize, our contributions in this study includes: 
\begin{enumerate}[label={(\alph*)}]
    \item We are the first, to the best of our knowledge, to define a backward compatibility metric for prediction explanation and propose BCX.
BCX utilizes practical agreement metrics to assess the consistency of explanations between old and new models.
    \item We propose BCXR, a BCX-aware retraining method that ensures theoretical validity by using differentiable surrogate losses to lower bound the non-differentiable agreement metrics.
    \item We conduct experiments on eight real-world datasets to validate the effectiveness of BCXR. The empirical evidence obtained from these experiments demonstrates the efficacy of our BCXR methods.
\end{enumerate}
    
% (a) We are the first, to the best of our knowledge, to define a backward compatibility metric for prediction explanation and propose BCX.
% BCX utilizes practical agreement metrics to assess the consistency of explanations between old and new models. (b) We propose BCXR, a BCX-aware retraining method that ensures theoretical validity by using differentiable surrogate losses to lower bound the non-differentiable agreement metrics. (c) We conduct experiments on eight real-world datasets to validate the effectiveness of BCXR.
% The empirical evidence obtained from these experiments demonstrates the efficacy of our BCXR methods.

The rest of the paper is organized as follows: We begin by introducing our notation and reviewing related works in Section 2.
Section 3 presents our proposed methods, BCX and BCXR.
In Section 4, we report the results of our numerical evaluation.
Finally, Section 5 concludes the paper.
The proofs of our theoretical analysis, the details of our experiments, and a discussion on the limitations of our method are provided in Appendix.

\section{Preliminary}
In this section, we briefly introduce the notation we use throughout this paper, as well as relevant previous methods.

\subsection{Notation}
We study supervised regression and classification problems.
The input space is $\scriptX \subseteq \doubleR^d$, where $\doubleR$ is the space of real values, $d \in \doubleN$ is the number of input features, and $\doubleN$ is the space of integers larger than zero.
The output space is $\scriptY \subseteq \doubleR$ for regression tasks and $\scriptY = [K]$ for classification tasks, where $[K]$ denotes the set of integers from 1 to $K \in \doubleN$, i.e., $[K]:=\{1,...,K\}$, and $K > 1$ is the number of classes.

We follow the model update schema with additional data, which is set up in studies of backward compatibility metrics~\citep{gbc2022sakai, abcd2023matsuno}.
Let $\scriptH =\{h: \scriptX \rightarrow \scriptY \}$ be a hypothesis space.
An old  model $h_1\in \scriptH$ is trained with data $D_1 := \{(\bx_i, y_i)\}_{i=1}^{n_1}$ drawn from a density denoted by $p(\bx,y)$ in an i.i.d. fashion.
After obtaining additional data $D_\Delta := \{(\bx_i, y_i)\}_{i=n_1 + 1}^{n_2} $ from $p(\bx,y)$, we train a new  model $h_2 \in \scriptH$ using $D_2 := D_1 \cup D_\Delta$.

An attributive explanation method $E: \scriptH \times \scriptX \rightarrow \doubleR^d$ provides the explanation of the prediction of a model $h \in \scriptH$ for an input $\bx \in \nset{X}$ by computing a vector of real values in $\doubleR^d$ whose $i$-th value represents the influences (e.g., importance, relevance, or contribution) of the $i$-th feature for the prediction $h(\bx)$.

\subsection{Related works}
Related works can be categorized into three groups: backward compatibility, explanation methods, and studies on disagreement in ML.

\subsubsection{Backward compatibility in ML}

The concept of backward compatibility in ML was originally introduced by Bansai et al.~\cite{btc2019bansai}, who proposed the \textit{Backward Trust Compatibility} (BTC) metric to measure the backward compatibility between old and new classification models ($h_1$ and $h_2$, respectively) as
\begin{align}
    \btc(h_1, h_2) := \frac{\doubleE_{p(\bx,y)} [\doubleI[h_1(\bx) = y \wedge h_2(\bx) = y]] }{\doubleE_{(\bx,y)} [\doubleI[h_1(\bx) = y]]},
\end{align}
where $\doubleI[P]$ represents the Iverson bracket, being 1 if the proposition $P$ is true and 0 otherwise, and $\doubleE_{p(\bx,y)}[f(\bx, y)] := \int_{\scriptX \times \scriptY} f(\bx,y ) p(\bx,y) dxdy$ denotes the expectation of $f(\bx,y)$ over the density $p(\bx,y)$.
BTC measures the ratio of correct predictions made by the new model among the samples for which the old model makes correct predictions. 
The authors then proposed a BTC-aware retraining objective for a new classifier $h_2$ defined by
\begin{align}
    L_{DM}(h_2) := \doubleE_{p(\bx,y)} \big[(1 + \lambda \doubleI[h_1(\bx) = y])\ell(h_2(\bx),y)\big] \eqlab{lbtc},
\end{align}
where $\ell:\scriptY \times \scriptY \rightarrow \doubleR{\geq 0}$ is a loss function, and $\lambda \in \doubleR_{>0}$ is a hyperparameter.
The minimization of \eqref{lbtc} is referred to as \textit{Dissonance Minimization} (DM)~\cite{btc2019bansai, gbc2022sakai}.
DM is versatile since it simply modifies the sample weights of the training data and hence it can be applied to most ML methods.

Another backward compatibility metric is \textit{Backward Error Compatibility} (BEC)~\cite{bec2020srivastava}, which focuses specifically on prediction errors.
Meanwhile, the \textit{Negative Flip Rate} (NFR)~\cite{nfr2021yan} counts the number of samples for which the old model makes correct predictions while the new model makes incorrect predictions.
Sakai~\cite{gbc2022sakai} generalized these backward compatibility metrics as a \textit{Generalized Backward Compatibility} (GBC) metric and theoretically established a generalization error bound of GBC-based learning.
Additionally, \textit{ABCD}~\cite{abcd2023matsuno} is proposed as a robust backward compatibility metric that defines compatibility based on the conditional distribution, which is approximated by $k$-nearest neighbors.
While a few backward-compatibility-aware retraining methods~\cite{nfr2021yan,gbc2022sakai, abcd2023matsuno} have been proposed beside DM, they are not as versatile as DM due to their objective customization.

\subsubsection{Explanation methods in ML}
Explainability is one of the most critical aspects of ML/AI systems, particularly in sensitive and critical domains such as healthcare and social security.
As a result, various eXplainable AI (XAI) methods have been proposed~\cite{xai2020, xai2023, iml}.
For fundamental tasks, intrinsically explainable methods, such as decision trees, linear models, and $k$-nearest neighbors, are utilized.
However, for complex tasks, black-box models like neural networks and kernel methods are commonly employed.
To explain these models, post-hoc feature-attribution-based explanation methods~\cite{fae2024wang} have been developed~\cite{lime2016, deeplift2017, anchors2018, gradcam2020, simonyan2014deep, integratedgrad2017sundararajan}.

One of the most prevalent explanation methods is SHAP~\cite{shap2017}, which utilizes the concept of Shapley values~\cite{shap1953} to explain a prediction by the sum of the contributions of each input feature.
While SHAP can be model-agnostic by implementing \textit{Kernel SHAP}~\cite{shap2017}, 
various specialized implementation have been proposed. 
For example, tree-based~\cite{treeshap2018}, gradient-based and other SHAP computation methods are officially available\footnote{\url{https://shap-lrjball.readthedocs.io/}}.
In addition, many SHAP-related research have been conducted for better approximation and faster computation~\cite{prishap2023, limitshap2021, AAS2021103502,imp2021,jethani2022fastshap, trackSHAP2022}.

%Conversely, SHAP can be efficient and precise for tree-based methods, including ensemble methods such as decision trees, random forests, and gradient-boosting tree regressors/classifiers~\cite{treeshap2018}.

\subsubsection{Disagreement measures of attributive explanations}
\seclab{disagreement}
It has been revealed that attributive explanations obtained from different methods often disagree with each other~\cite{order2021, disagree2022krishna}.
To measure the disagreements between two explanation methods for a single model, various metrics have been proposed~\cite{order2021, disagree2022krishna, flora2022comparing1,  flora2022comparing2}.
For example, \citet{disagree2022krishna} proposed top-$k$ feature agreement (S{\o}rensen–Dice coefficient of top-$k$ features), top-$k$ rank agreement, top-$k$ sign agreement, top-$k$ signed rank agreement, based on practitioners' perspectives.
Since practically meaningful agreement metrics may depend on applications, these various design of metrics are important.
The agreement measures are defined as follows;
\begin{align}
    &\featagree(\be_1, \be_2; k) \notag \\ 
    &:= \frac{1}{k}\Big| \big\{ i \in [d] ~\big|~ i \in \topfeat(\be_1;k) \wedge i \in \topfeat(\be_2;k) \big\}  \Big|
\end{align}
\begin{align}
    &\rankagree(\be_1, \be_2; k) \notag \\ 
    &:= \frac{1}{k}\Big| \big\{ i \in [d] ~\big|~ i \in \topfeat(\be_1;k) \wedge i \in \topfeat(\be_2;k) \notag \\
    & \hspace{3em} \wedge  \rank(\be_1, i) = \rank(\be_2, i) \big\}  \Big|\\
    &\signagree(\be_1, \be_2; k) \notag \\ 
    &:= \frac{1}{k}\Big| \big\{ i \in [d] ~\big|~ i \in \topfeat(\be_1;k) \wedge i \in \topfeat(\be_2;k) \notag \\
    & \hspace{3em} \wedge  \sign(e_{1i}) = \sign(e_{2i}) \big\}  \Big| \\
    &\signedrankagree(\be_1, \be_2; k) \notag \\ 
    &:= \frac{1}{k}\Big| \big\{ i \in [d] ~\big|~ i \in \topfeat(\be_1;k) \wedge i \in \topfeat(\be_2;k) \notag \\
    & \hspace{3em} \wedge  \sign(e_{1i}) = \sign(e_{2i}) \wedge \rank(\be_1, i) = \rank(\be_2, i)  \big\}  \Big|
\end{align}
where 
$\rank(\bx, i) := |\{j \in [d] \mid |x_j| \geq |x_i| \}|$ outputs the rank of the absolute of $x_i$ among the absolutes of elements of $\bx$ in \textit{descending} order (i.e., $|x_i|$ is the ($\rank(\bx, i)$)-th largest value among $|x_1|,...,|x_d|$)\footnote{When $\exists i > j \in [d], 
 |x_i| = |x_j|$, we set $\rank(\bx, j) = \rank(\bx, i) + 1$ for consistency.},
$\topfeat(\bx; k) := \{ i \in [d] \mid \rank(\bx, i) \leq k \} $ is the set of indices where that ranks of the corresponding elements of $\bx$ are smaller than or equal to $k$ (i.e., set of indices of features whose absolute value is at least $k$-th largest),
and $\sign(x) := 1$ if $x\geq 0 $ else $-1$, is the sign of $x$.\footnote{We abuse to define $\sign(0) = 1$ for mathematical simplicity in our theoretical analysis.}
Note that these agreement metrics are invariant to the replacement of $\be_1$ and $\be_2$.

Although our interest aligns with these studies to some extent and we utilize the agreement metrics exemplified above, we aim at investigating the differences between \textit{two models} using a \textit{single explanation method}, where differences between \textit{two explanation methods} for \textit{a single model} have been studied.
Thus, although Neely et al.~\cite{order2021} conclude that agreement is not a suitable criterion for evaluating explanations, we still maintain that the explanations of both old and new models should agree for consistent model updates.

\section{Proposed method}
In this section, we first propose our Backward Compatibility metric in eXplanations, which we call BCX.
Then we present our BCX-aware Retraining method, which we call BCXR.
Please note that we omit $k$ from notation of agreement metrics in our analyses, e.g., we denote $\agree(\be_1, \be_2)$ instead of $\agree(\be_1, \be_2; k)$ for the sake of readability, while the statements hold true for any choice of $k \in [d]$.
In addition, all proofs are presented in our appendix.
%In addition, proofs of all lemmas are presented in our appendix (supplementary material) due to the space limitation.

\subsection{Backward compatibility in explanations}
We define the backward compatibility metric in terms of attributive explanation of models' prediction as follows using any choice of explanation method and agreement metric to quantify the agreement between two explanations.

\begin{definition}[Backward Compatibility in eXplanations]
    \deflab{bcx}
    Given two models $h_1$ and $h_2\in \scriptH$,
    % an evaluation data $\Dev = \{(\bx_i, y_i)\}_{i=1}^{n}$,
    an attributive explanation method $E:\scriptH \times \scriptX \rightarrow \doubleR^d$,
    and an agreement metric $\agree: \doubleR^d \times \doubleR^d \rightarrow [0,1]$,
    the Backward Compatibility in eXplanation (BCX) of $h_2$ over $h_1$ is defined as
    \begin{align}
        &{\bcex}(h_1, h_2; \agree, E) \notag \\
        &:= \frac{
            \doubleE_{p(\bx, y)} \left[ \agree(E(h_1, \bx), E(h_2, \bx))  \cdot s(\bx, y; h_1, h_2) \right]
        }{
            \doubleE_{p(\bx, y)} \left[  s(\bx, y; h_1, h_2) \right]
        }  ,
    \eqlab{empbcex}
\end{align}
where the sample selection function $s(\bx, y; h_1, h_2)$ is defined as
\begin{align}
    s(\bx, y; h_1, h_2) &:= c(h_1(\bx), y) \cdot c(h_2(\bx), y)
\end{align}
and 
\begin{align}
    c(\widehat{y}, y)  &:=\begin{cases}
        \doubleI\left[(\widehat{y} - y)^2 \leq \tau \right] & \text{(regression)}\\
        \doubleI\left[\widehat{y} = y\right] & \text{(classification)}
    \end{cases}
    \eqlab{correct}
\end{align}
is the correctness of the prediction by $h$ for the sample $(\bx, y)$.
$\tau$ is a predefined threshold to determine the correctness for regression tasks.%
\footnote{$\tau$ can be a user-defined hyperparameter. For example, we set the threshold $\tau$ to be the empirical mean squared error (MSE) of an old model $h_1$ (i.e., we use $\tau :=  1/|D_2|\sum_{(\bx, y) \in D_2}(h_1(\bx) - y)^2$ in our experiments).} 
\end{definition}

Our definition of BCX is both practical and meaningful.
A straightforward approach to defining BCX involves computing the expected agreement scores between  $h_1$ and $h_2$, for all samples $(\bx, y) \sim p(\bx, y)$.
For instance, this can be achieved by computing $\doubleE_{p(\bx, y)} [ \agree(E(h_1, \bx), E(h_2, \bx))  ]$.
However, this approach may not be suitable for practical use due to two reasons.
First, aligning explanations when the old model gives incorrect predictions (e.g., $h_1(\bx) \neq y$) may have little practical value.
Second, it is impractical to have aligned explanations when the new model provides incorrect predictions (e.g., $h_2(\bx) \neq y$).
Hence, it is essential to focus on the agreement scores for samples where both old and new models make correct predictions.
Based on this motivation, we have devised our definition of BCX.
A high BCX score indicates that the new model consistently provides compatible explanations for samples with compatibly correct prediction.

% We specifically utilize SHAP for the explanation method $E$ due to 
In this work, we mainly investigate SHAP for the explanation method $E$ due to its prevalence in real applications and recent active studies.
To assess the agreement between explanations, we employ the four agreement metrics introduced in \secref{disagreement}.
These metrics are specifically designed from a practitioner's perspective and offer practical utility.

% BCX can utilize any attributive explanation method,

% Previous studies have reported that different explanation methods often yield disagreements~\cite{disagree2022krishna}, indicating that the score of BCX can potentially be influenced by the choice of the explanation method $E$.
% In this work, we mainly investigate SHAP for $E$ due to its prevalence in real applications and recent active studies.
% Specifically, we utilize the gradient-based SHAP method due to its differentiability, which is crucial for BCX-aware retraining.
% It is important to note that the explanation method $E$ can be any as long as it is differentiable w.r.t. its inputs.
% For instance, gradient-based explanation methods~\cite{deeplift2017,gradcam2020} can also be used for our BCX-aware retraining method, explained in the next section.
% % The details of BCX-aware retraining will be explained in the next section.

% To assess the agreement between explanations, we employ the four agreement metrics introduced in \secref{disagreement}.
% These metrics are specifically designed from a practitioner's perspective and offer practical utility.
% % It is important to note that we set the threshold $\tau$ to be the empirical mean squared error (MSE) of an old model $h_1$ (i.e., we use $\tau :=  1/|D_2|\sum_{(\bx, y) \in D_2}(h_1(\bx) - y)^2$ in our experiments).

\subsection{BCX-aware retraining}
% \begin{ldefinition}
%     Given $h_1 \in \scriptH$, the problem is to find $h_2 \in \scriptH$ which shows a high predictive performance as well as a high score of $ \bcx(h_1, h_2; \agree, E) $.
% \end{ldefinition}

Next, we aim at training a new model $h_2$ where a high BCX score of $h_2$ over $h_1$ is preferred.
The training objective of $h_2$ is naturally formulated as follows, similarly with the formulation in \citep{btc2019bansai, abcd2023matsuno}.
\begin{align}
\eqlab{rh2}
    &R(h_2) := \doubleE_{p(\bx, y)}[\ell(h_2(\bx), y)] + \lambda (1- \bcx (h_1, h_2; \agree, E)),
\end{align}
where $\ell: \scriptY \times \scriptY \rightarrow \doubleR_{\geq 0}$ is a loss function, e.g., squared error for regression and 0-1 loss for classification.
$\agree$ is one of feature-agreement,  rank-agreement, sign-agreement, and signedrank-agreement with given $k$.

\block{Differential surrogate loss design.}
In order to train the model $h_2$, it is necessary for its objective function to be differentiable w.r.t. the parameters of $h_2$.
Then, we have two issues regarding the differentiability;
The one is the differentiability of the explanation method $E$ and the other is the differentiability of the agreement metric $\agree$.
For the former, we can use a differentiable explanation method for $E$ and use a differentiable model for $h_2$ (e.g., neural networks).
Specifically, we use the gradient-based SHAP as a differentiable SHAP computation in our experiments.
It should be noted that our formulation and analysis are general and hence other differentiable explanation methods~\cite{deeplift2017,gradcam2020,simonyan2014deep, integratedgrad2017sundararajan} can also be utilized.
%(e.g., the gradient-based SHAP) and a differentiable model $h$ (e.g., neural networks).

For the latter, however, the agreement metrics lack differentiability due to their discrete nature.
% We use a differentiable explanation method for $E$ (e.g., the gradient-based SHAP) and a differentiable model $h$ (e.g., neural networks).
% However, the agreement metrics lack differentiability due to their discrete nature.
% Despite our goal of minimizing $R(h_2)$, we cannot use it directly as the objective due to its lack of differentiability caused by the discrete nature of the agreement metrics.
% Note that we use a differentiable explanation method for $E$ (e.g., the gradient-based SHAP) and a differentiable model $h$ (e.g., neural networks) for the differentiability beside agreement metrics.
% Note that the explanation method $E$, specifically the gradient-based SHAP, can be made differentiable by representing $h_2$ with neural networks.
Consequently, we propose a differentiable surrogate loss that provides an upper bound for $1- \bcx (h_1, h_2; \agree, E)$ in equation \eqref{rh2}, in order to design a differentiable objective.
Specifically, we first consider feature-agreement and we define our surrogate loss $\lfeat(\be_2; \be_1, k)$ to lower bound $\featagree(\be_1, \be_2; k)$ as follows
\begin{align}
    &\lfeat(\be_2; \be_1, k)  \notag \\
    &:= \frac{1}{k} \sum_{i\in \topfeat(\be_1;k)} \max\big(0, \psi_{feat}(\be_2) - |e_{2i}| + \varepsilon \big),
\end{align}
where $\psi_{feat}: \doubleR^d \rightarrow \doubleR$ is defined as 
\begin{align}
    \psi_{feat}(\be_2) := 
    \begin{cases}
        \max_{i \not\in \topfeat(\be_1;k) } |e_{2i}| & (k < d) \\
        -\varepsilon & (\text{otherwise})
    \end{cases}
\end{align}
and 
$\varepsilon > 0$ is a predefined small constant.
%where $\varepsilon > 0$ is a predefined small constant.
We establish the following lemma between the surrogate loss $\lfeat$ and feature-agreement metric $\featagree$, which provides theoretical validity of the use of $\lfeat$.
\begin{lemma}
\lemlab{feat}
% For any $\be_1, \be_2$, $1 - \featagree(\be_1, \be_2; k) \leq \varepsilon^{-1} \lfeat(\be_2; \be_1, k)$ holds.
The following inequality holds for any $\be_1, \be_2$, and $k$.
\begin{align}
    1 - \featagree(\be_1, \be_2; k) \leq \varepsilon^{-1} \lfeat(\be_2; \be_1, k) %\eqlab{ntjg}
\end{align}
\end{lemma}
\lemref{feat} shows that $\lfeat$ multiplied with $\varepsilon^{-1}$ upper bounds one minus feature-agreement (i.e., feature-\textbf{dis}agreement) and hence minimization of $\lfeat$ maximizes the score of feature-agreement.

\block{Objective for BCXR.}
Now we can upper bounds the non-differentiable term $(1 - BCX(h_1, h_2; \featagree, E))$ in \eqref{rh2} based on \lemref{feat} as
\begin{align}
    &1 - BCX(h_1, h_2; \featagree, E) \notag\\
    &= \frac{
            \doubleE_{p(\bx, y)} \big[ \big(1 - \featagree(E(h_1, \bx), E(h_2, \bx))\big)   s(\bx, y; h_1, h_2) \big]
        }{
            \doubleE_{p(\bx, y)} \left[  s(\bx, y; h_1, h_2) \right]
        }\\
    &\leq  \varepsilon^{-1} \frac{
            \doubleE_{p(\bx, y)} \big[\lfeat(E(h_2, \bx); E(h_1, \bx), k)   s(\bx, y; h_1, h_2) \big]
        }{
            \doubleE_{p(\bx, y)} \left[  s(\bx, y; h_1, h_2) \right]
        }.
\end{align}
Hence we have the following upper bound of $R(h_2)$ with $\featagree$, which is differentiable w.r.t. $h_2$;
\begin{align}
    R(h_2) \leq&  \doubleE_{p(\bx, y)}[\ell(h_2(\bx), y)] \notag \\
    &  + \lambda  \frac{
            \doubleE_{p(\bx, y)} \big[\lfeat(E(h_2, \bx); E(h_1, \bx), k)   s(\bx, y; h_1, h_2) \big]
        }{
            \doubleE_{p(\bx, y)} \left[  s(\bx, y; h_1, h_2) \right]
        } \eqlab{asdfas}\\
    =:& ~\Lfeat(h_2),
\end{align}
where the constant $\varepsilon^{-1}$ is absorbed by $\lambda$ for simplicity and we denote the right hand of \eqref{asdfas} by $\Lfeat(h_2)$.
In practical scenarios, we resort to the empirical approximation of $\Lfeat(h_2)$ for the BCX-aware retraining, since we cannot know  the underlying distribution $p(\bx,y)$.
Formally, our proposed feature-agreement-based BCX-aware retraining method (referred to as BCXR-Ftr) is defined as follows.
\begin{definition}[Feature-agreement-based BCX-aware Retraining (BCXR-Ftr)]
\deflab{BCXR}
Given an old model $h_1:\scriptX \rightarrow \scriptY$,
and a training data $D := \{(\bx_i, y_i)\}_{i=1}^n$,
BCXR trains a new model $h_2$ by minimizing the following objective;
\begin{align}
        &\widehat{L}_{\mathrm{Ftr}}(h_2;D)
    := \frac{1}{|D|}\sum_{(\bx, y) \in D} \ell(h_2(\bx), y) \notag \\
    &\quad + \lambda \frac{1}{|D_s|}\sum_{\bx \in D_s} \lfeat(E(h_2, \bx); E(h_1, \bx),k) \eqlab{empobjective}
\end{align}
where
$D_s := \{\bx \mid (\bx, y) \in D \wedge s(\bx, y; h_1, h_2) = 1\}$ is the set of samples where $h_1$ and $h_2$ make correct predictions,
and $\lambda \in \doubleR_{\geq 0}$ is a hyperparameter.
\end{definition}

Similar surrogate losses for other agreement metrics, i.e., rank-, sign-, and signedrank-agreements are defined to lower bound each of agreements with theoretical analyses in the following lemmas from \lemref{lrank} to \lemref{lsignedrank}.
\begin{lemma}
\lemlab{lrank}
Let $ I = \{(j, {\argsort(\abso(\be_1))}_j) \mid j \in [k] \}$ be the set of tuples each $(j,i)$ of which indicates that the $i$-th element of $\be_1$ is the $j$-th largest value among $\{|e_{11}|,..., |e_{1d}|\}$, where $\argsort$ returns the indices that would sort its input in \textit{descending} order.
Then we define $\lrank$ as 
\begin{align}
    &\lrank(\be_2; \be_1, k)\notag \\ 
    & : = \frac{1}{k} \sum_{(j,i) \in I}  \max \bigg(0,  \sort(\abso(\be_2)_{i=-\varepsilon})_j - |e_{2i}| + \varepsilon \bigg)
\end{align}
where $\ba_{i=x} := [a_{1},...,a_{i-1},x,a_{i+1},...,a_{d}] $ is a copy of $\ba$, whose $i$-th element is replaced to $x$. 
Then we have following inequality for any $\be_1, \be_2$ and $k$,
\begin{align}
    1 - \rankagree(\be_1, \be_2; k) \leq \varepsilon^{-1} \lrank(\be_2; \be_1, k)
\end{align}
where $\sort$ sorts its input in \textit{descending} order.
\end{lemma}
\begin{lemma}
\lemlab{lsign}
Let us define $\lsign$ as
\begin{align}
    &\lsign(\be_2; \be_1, k) \notag \\ 
    & : = \frac{1}{k} \sum_{i\in \topfeat(\be_1;k)} \max\big(0, \psi_{sign}(\be_2) - \sign(e_{1i})e_{2i} + \varepsilon \big)
\end{align}
where
\begin{align}
    \psi_{sign}(\be_2) :=
    \begin{cases}
        \max_{i \not\in \topfeat(\be_1;k) } |e_{2i}| & (k < d) \\
        0 & (\text{Otherwise})
    \end{cases}.
    \end{align}
Then we have following inequality for any $\be_1, \be_2$ and $k$,
\begin{align}
    1 - \signagree(\be_1, \be_2; k) \leq \varepsilon^{-1} \lsign(\be_2; \be_1, k)
\end{align}
\end{lemma}

\begin{lemma}
\lemlab{lsignedrank}
Let us define $\lsignedrank$ with $I$ defined in \lemref{lrank} as
\begin{align}
    &\lsignedrank(\be_2; \be_1, k)  \\ 
    &:= \frac{1}{k} \sum_{(j,i) \in I}  \max \bigg(0,  \sort(\abso(\be_2)_{i=0})_j - \sign(e_{1i})e_{2i} + \varepsilon \bigg).
\end{align}
Then we have following inequality for any $\be_1, \be_2$ and $k$,
\begin{align}
    1 - \signedrankagree(\be_1, \be_2; k) \leq \varepsilon^{-1} \lsignedrank(\be_2; \be_1, k)
\end{align}
\end{lemma}

Based on these lemmas, we define rank-agreement-, sign-agreement-, and signedrank-agreement-based BCXR (denoted by BCXR-Rnk, BCXR-Sgn, and BCXR-SgnRnk, respectively) with corresponding objectives $\widehat{L}_{\mathrm{Rnk}}, \widehat{L}_{\mathrm{Sng}}$ and $\widehat{L}_{\mathrm{SgnRnk}}$ as similar with feature-based BCXR, defined in \defref{BCXR}, by replacing $\lfeat$ in \eqref{empobjective} by $\lrank, \lsign$  and $\lsignedrank$, respectively.
Each of these are specialized objective to specifically improve the corresponding agreement metric.

\subsection{Universal BCXR}
\seclab{universal}
We have devised surrogate loss functions to enhance the four agreement metrics.
Furthermore, we propose a universal loss function that can effectively lower bound all agreement metrics regardless of the choice of $k$.
Specifically, we utilize $\normdisagree$, which represents the Euclidean distance between the explanations, defined as follows:
\begin{align}
    \normdisagree(\be_1, \be_2) := \enorm{\be_1 - \be_2} = \sqrt{\sum_{i=1}^d (e_{1i} - e_{2i})^2 }.
\end{align}
$\normdisagree$ is differentiable and can be used as a loss function directly.
For the validity of the use of $\normdisagree$, we establish the following lemma.
\begin{lemma}
\lemlab{ansdjfj}
Given any $\be_1 \in \doubleR^d$ and $k \in [d]$,
assume that there exists $\delta > 0$ such that (a)
 $\bigl||e_{1i}| - |e_{1j}|\bigr| \geq \sqrt{2}\delta$ holds for any $i \neq j \in \mathrm{TopFeatures}(\be_1, \max(k+1, d))$, and 
(b) additionally if $k = d$, $ |e_{1i}| \geq \delta $ holds for any $i \in [d]$.\footnote{These assumptions may be satisfied by a proper feature engineering,.e.g, both features with the same effects to the prediction and ones with constantly zero effects to the prediction can be removed from features.}
Then the following inequality holds for any $\be_2 \in \doubleR^d$.
\begin{align}
    1 - \signedrankagree(e_1, e_2, k) \leq \delta^{-1} \normdisagree(\be_1, \be_2)
\end{align}
\end{lemma}
Note it is trivial by definition that feature agreement lower bounds both rank agreement and sign agreement, each of which lower bounds signed-rank agreement, i.e.,  the following inequality holds true for any $\be_1, \be_2$ and $k$ as
\begin{align}
    \eqlab{agreementorder}
    \featagree(\be_1, \be_2)  \geq \begin{array}{c}
        \rankagree(\be_1, \be_2)  \\
        \signagree(\be_1, \be_2)
    \end{array}\geq \signedrankagree(\be_1, \be_2).
\end{align}
By \eqref{agreementorder} and \lemref{ansdjfj},
$\normdisagree$ multiplied with a constant $\delta^{-1}$ upper bounds any disagreement, i.e., one minus feat-, sign-, rank-, and signedrank-agreement for any choice of $k$.
Hence, regardless of which agreement metric is used to calculate BCX, $\normdisagree$ can be used for the loss function responsible for the compatibility of BCX.
Furthermore, the use of $\normdisagree$ eliminates the burden to tune the hyperparameter $\varepsilon$.

\block{Universal BCXR objective.}
Finally, we define our universal objective for our BCXR, which replaces $\lfeat$ in \eqref{empobjective} by $\normdisagree$ as follows.
\begin{align}
    &\widehat{L}_{\mathrm{Norm}}(h_2;D)
    := \frac{1}{|D|}\sum_{(\bx, y) \in D} \ell(h_2(\bx), y) \notag \\
    &\quad + \lambda \frac{1}{|D_s|}\sum_{(\bx, y) \in D_s} \normdisagree(E(h_1, \bx), E(h_2, \bx))
    \eqlab{BCXRnorm}
\end{align}
We refer to the minimization of \eqref{BCXRnorm} as BCXR-Norm.

We have established all of our BCXR methods and next investigate the empirical behaviour of BCXR for real-world data sets.

\begin{table}[t]
\caption{Data set statistics.}
\label{tab:table}
\begin{center}
\begin{small}
% \begin{sc}
\setlength\tabcolsep{6pt}
%setlength\extrarowheight{2pt}
%\renewcommand{\arraystretch}{1.07}
\begin{tabular}{clrr} \toprule
Task&Data set  &Samples&Features\\ \midrule
%&australian&690    &14      \\
%&german    &1000   &24      \\
%&svmguide3 &1243   &22      \\
\multirow{4}{*}{\rotatebox[origin=c]{0}{Regresssion}}
&space-ga          &3107   &6       \\
% &abalone          &4177   &8       \\
&cadata           &20640  &8       \\
&cpusmall         &8192   &12      \\
&YearPredictionMSD&463715 &90      \\
\midrule
\multirow{5}{*}{\rotatebox[origin=c]{0}{Classification}}
% &svmguide1 &3089   &4       \\
&cod-rna   &59535  &8       \\ 
% &ijcnn1    &49990  &22      \\
&phishing  &11055  &68      \\
% &mushrooms &8124   &112     \\
&a9a       &32561  &123     \\
&w8a       &49749  &300     \\
\bottomrule
\end{tabular}
% \end{sc}
\end{small}
\end{center}
\end{table}

\begin{figure*}[t]
    \begin{center}
    % \centering
    \subfigure[space-ga]{\includegraphics[width=\textwidth]{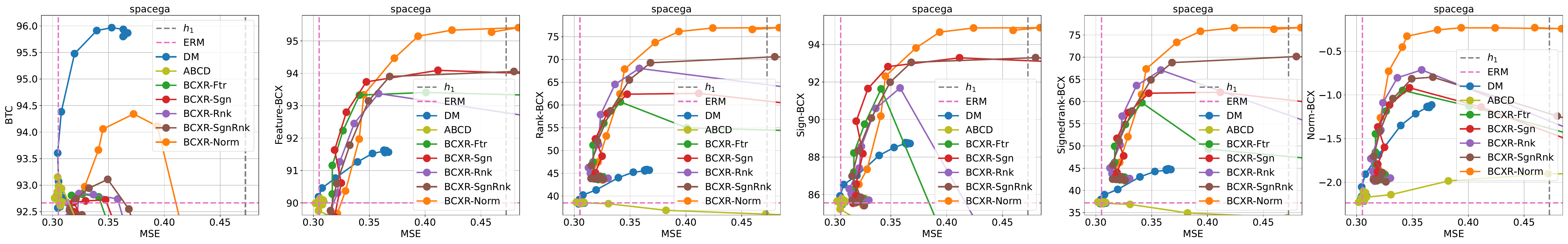}}
    % \subfigure[abalone]{\includegraphics[width=\textwidth]{fig/tradeoff-abalone.pdf}}
    \subfigure[cadata]{\includegraphics[width=\textwidth]{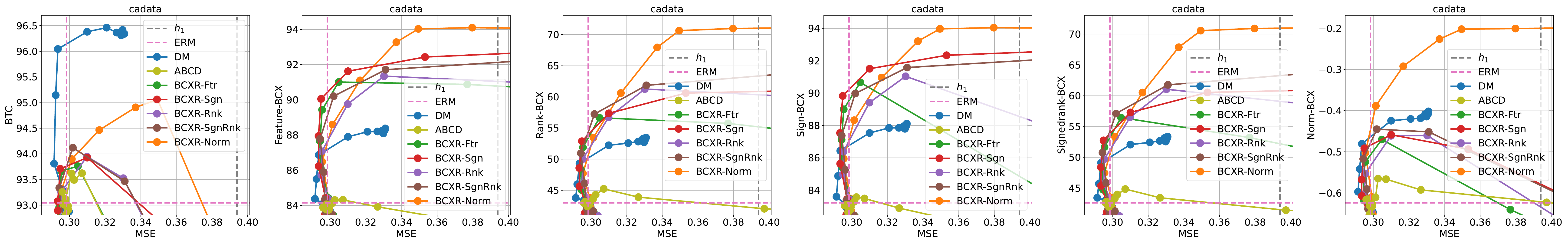}}
    \subfigure[cpusmall]{\includegraphics[width=\textwidth]{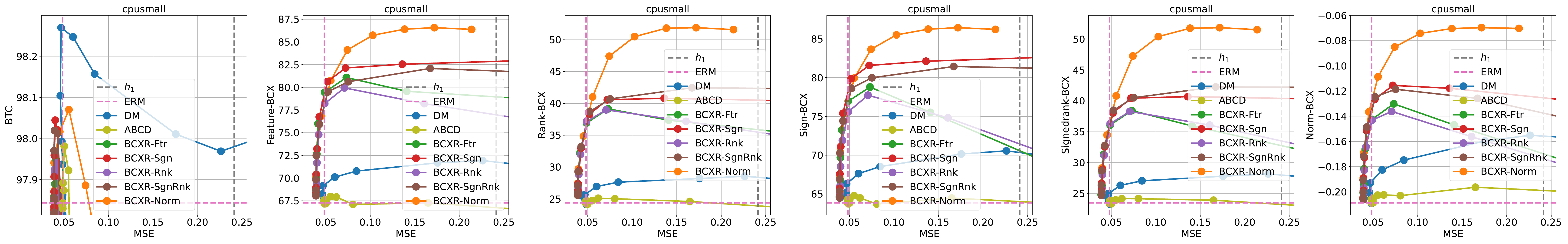}}
    \subfigure[YearPredictionMSD]{\includegraphics[width=\textwidth]{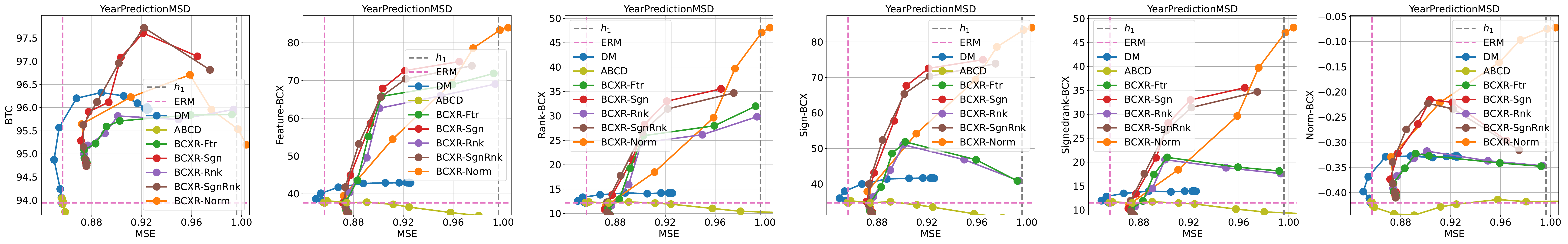}}
    \caption{
    Trade-off for regression data sets.
Horizontal axes represents MSE (the lower the better, $\leftarrow$) and vertical axes represents each of BTC and BCXs with different agreement metrics (the higher the better, $\uparrow$).
In general, points located in the upper left region of each figure indicate better results compared to points in the lower right region.
The grey dashed vertical lines indicate the MSE achieved by old models.
The pink dashed vertical and horizontal lines represent the MSE and backward compatibility scores achieved by the ERM.
Retraining methods that take backward compatibility into account are expected to perform better MSE than the old models (up to the grey dashed lines) and better compatibility than ERM  (up to the pink horizontal lines).
Since this is a multi-objective optimization problem, the results on the Pareto fronts are considered effective in finding better trade-offs between MSE and backward compatibility scores.
    }
    \label{fig:tradeoff-regression}
    \end{center}
\end{figure*}

\begin{figure*}[t]
    \begin{center}
    % \centering
    \subfigure[cod-rna]{\includegraphics[width=\textwidth]{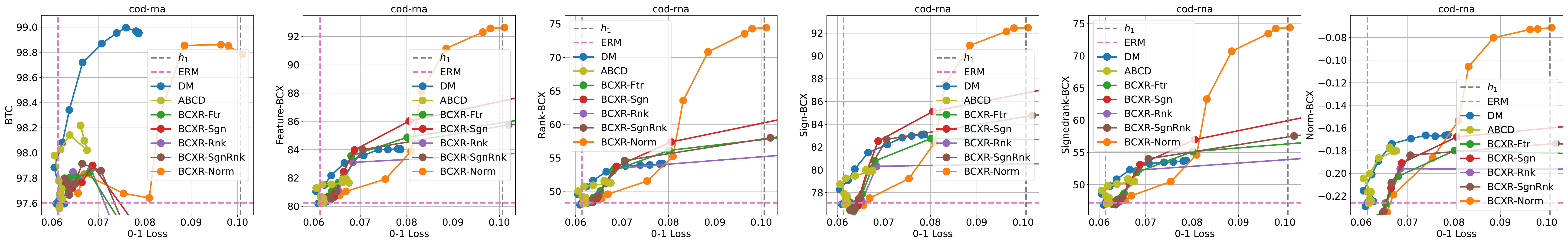}}
    % \subfigure[ijcnn1]{\includegraphics[width=\textwidth]{fig/tradeoff-ijcnn1.pdf}}
    \subfigure[phishing]{\includegraphics[width=\textwidth]{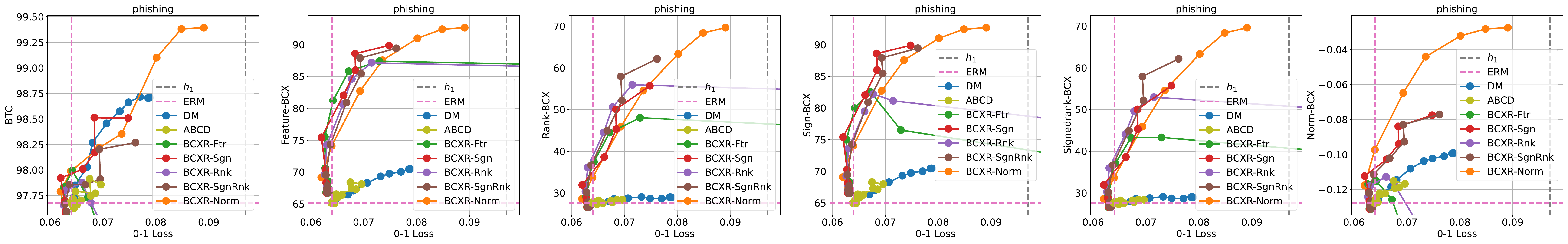}}
    \subfigure[a9a]{\includegraphics[width=\textwidth]{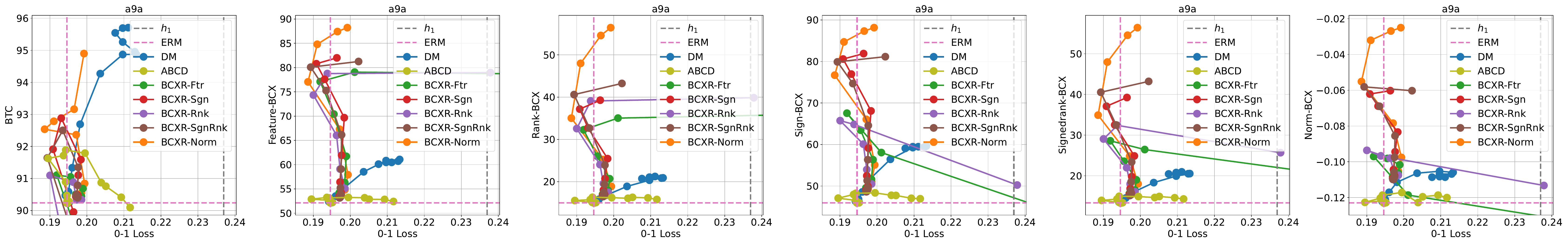}}
    \subfigure[w8a]{\includegraphics[width=\textwidth]{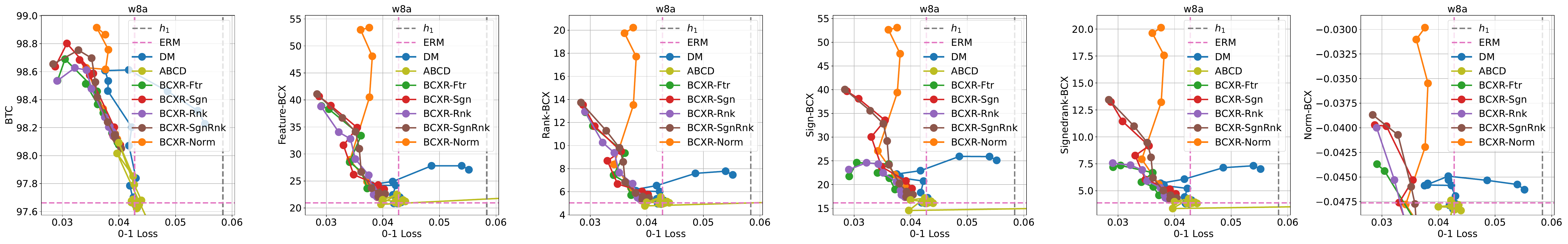}}
    \caption{Trade-off for classification data sets. Explanation of figures follow \figref{tradeoff-regression}.}
    \label{fig:tradeoff-classification}
    \end{center}
\end{figure*}

\section{Experiments}
We conduct experiments on real world data sets to verify the effectivity of the proposed objective function.
The implementation is based on python with PyTorch~\citep{pytorch2019paszke} and scikit-learn~\citep{sklearn}.
All experiments are carried out on a computational server equipping four Intel Xeon Platinum 8260 CPUs with 192 logical cores in total and 1TB RAM.

%\block{Data.}
\subsection{Data set}
We utilize four regression and four classification data sets obtained from LIBSVM~\cite{libsvm}.
We vary the number of input features from 6 to 300 in order to study the behavior of BCXR w.r.t. $d$.
The data set statistics are presented in \tabref{table}.

\subsection{Setting}
We randomly sample 2000 samples from each data set.
The first 200 samples are used as $D_1$ and the first 1000 samples including $D_1$ are used as $D_2$.
The remaining 1000 samples are used for evaluation.

We utilize three-layer neural networks to model old and new models.
The number of hidden units are 100,
the activation function is ReLU~\cite{relu2010Nair},
and we use batch normalization~\cite{batchnorm2015Ioffe} after each of the activation layers.
Each of the $D_1$ and $D_2$ is split into training and validation sets at an ratio of $80:20$ and the validation set are used for early stopping~\cite{early1989Morgan}.
The Adam optimizer with a learning rate of 0.01 and weight decay~\cite{weightdecay1988hanson} of $1 \times 10^{-4}$ are used for training.
The maximum number of epochs is set to 200.

The old model $h_1$ is trained by a standard model training method, i.e., empirical risk minimization (ERM) with $D_1$.
New models are trained with our BCXR-Ftr, BCXR-Rnk, BCXR-Sgn, BCXR-SgnRnk, and BCXR-Norm.
The comparison baselines are ERM, DM (BTC-aware retraining method~\cite{btc2019bansai}), and ABCD (ABCD-aware retraining method~\cite{abcd2023matsuno}).
% BCXR methods using $\lfeat$, $\lrank$, $\lsign$, $\lsignedrank$ and $\normdisagree$ as well as  ERM and DM (dissonance minimization~\cite{btc2019bansai}) 
% as baselines.
For DM and ABCD we vary their hyperparameter $\lambda$ from $1 \times 10^{-4}$ to $1 \times 10^{4}$.
Other hyperparameters for ABCD follow its author-defined values.

For our BCXR methods, we set $\varepsilon = 1 \times 10^{-3}$,
% and $k=5$ for $\lfeat, \lrank, \lsign$ and $\lsignedrank$
and for regression tasks, we set the threshold $\tau$ to $ 1/|D_2|\sum_{(\bx, y) \in D_2}(h_1(\bx) - y)^2$.
The hyperparameter $\lambda$ is set from $1 \times 10^{-4}$ to $1\times 10^{2}$.

The evaluation metrics are the standard loss (i.e., mean squared error (MSE) for regression tasks and mean 0-1 loss for classification tasks), BTC and empirical BCX scores with feature-, rank-, sign-,  signedrank-, and norm-agreement metrics on evaluation data\footnote{Since $\normdisagree$ is a \textbf{dis}agreement metric, we use norm-agreement as $1-\normdisagree(\be_1, \be_2)$, and we use it in \eqref{empbcex} for our evaluation.}.
The value of $k$ set to 5 following the existing study~\cite{disagree2022krishna}, and the same $k$ is used for our BCXR methods as well.
We repeat our experiments for 30 times for each data set and each methods with each settings of $\lambda$, and report the average scores for each of the evaluation metrics.

\subsection{Results}
The results are presented in \figref{tradeoff-regression} for regression datasets and \figref{tradeoff-classification} for classification datasets, where the backward compatibility scores are plotted against the loss (MSE and 0-1 loss).
The explanation of the plots are as follows;
the vertical grey dashed lines indicate the loss values of the old models.
The intersections of the horizontal and vertical pink dashed lines represent the losses and backward compatibility scores of ERM.
Therefore, backward-compatibility-aware retraining methods are expected to be positioned above the horizontal pink dashed lines and to the left of the vertical grey dashed lines.
Each of our methods and baselines follows a bottom-to-top pattern as the parameter $\lambda$ increases.
For example, the result with a value of $\lambda = 1\times 10^{-4}$ is located near the pink dashed lines, while results with larger $\lambda$ values are found in the upper parts of each figure.
Since there are often trade-offs between the loss and BTC and BCX metrics, and the importance of compatibility varies depending on the application, it is difficult to determine the best retraining method and parameter value of $\lambda$.
However, the methods that form Pareto fronts in the figures are generally considered to be effective.
We now discuss the details of the regression and classification results.

%\block{Results for regression tasks.}
\subsubsection{Results for regression tasks.}
The BCXR-Norm method forms part of the Pareto fronts in all BCX-based plots for regression data sets, demonstrating the universality of BCXR.
This result confirms the findings discussed in \secref{universal}.
Additionally, we observe that for results with smaller MSE, other BCXR methods (BCXR-Ftr, -Rnk, -Sgn, and BCXR-SgnRnk) offer comparable or better trade-offs than BCXR-Norm.
This is particularly evident in the case of YearPredictionMSD, as $\normdisagree$ considers all 90 features, while only the top 5 features significantly influence the agreement scores.
As a result, BCXR methods other than BCXR-Norm effectively consider these five features and consequently optimize new models more efficiently.
Furthermore, among the BCXR results excluding BCXR-Norm, both BCXR-Sgn and BCXR-SgnRnk consistently demonstrate better trade-offs in all BCX scores compared to BCXR-Ftr and BCXR-Rnk.
This suggests that enforcing constraints based on the signs of the explanations is crucial for maintaining consistent explanations of the top-$k$ features.

Among the baselines, DM exhibits some improvements in the BCX scores as $\lambda$ increases, and it partially contributes to the formation of the front lines.
However, the degree of improvement is relatively limited.
Interestingly, in the case of YearPredictionMSD, our BCXR methods outperform DM in terms of the BTC score.
Additionally, since ABCD focuses on enhancing conditional losses, it does not exhibit better trade-offs in our evaluation.
These findings further underscore the superiority of BCXR in providing consistent explanations during model updates.

\subsubsection{Results for classification tasks.}
The results obtained from the cod-rna and phishing exhibit similar patterns to those observed in regression tasks.
Specifically, when the 0-1 losses are small, agreement-based BCXR methods yield better results.
Conversely, BCXR-Norm demonstrates better trade-offs when the losses are large.
Interestingly, for a9a and w8a, datasets with large numbers of features (123 and 300, respectively), which are significantly larger than the value of $k=5$, our BCXR methods not only improve the BCX scores but also enhance BTC and reduce 0-1 losses.
These outcomes suggest that the inclusion of BCX as a constraint potentially leads to the improved optimization of training new models.
Consequently, our BCXR methods are proven to be effective and can be applied across a wider range of applications, extending beyond the sole purpose of maintaining explanation compatibility.

% Overall, we confirmed that 
% (a) agreement-based BCXR, in particular BCXR-Sgn and BCXR-SgnRnk, are great for BCX-aware retraining, forming Pareto front lines in loss vs BCX plots,
% (b) BCXR-Norm often improves all of BCX scores, demonstrating its versatility,
% and (c) BCXR even improves BTC and accuracy for classification data sets with large number of features.
% These results indicate the effectiveness of our proposed BCXR methods.

\section{Conclusion}
In this study, we have introduced BCX as a novel approach to assess the consistency of explanations in model updates.
Then, to overcome the challenge of non-differentiability in the agreement metrics, we propose differential surrogate losses that possess theoretical validity for substitution.
Building upon this, we have proposed BCXR, a BCX-aware retraining method, which leverages the surrogate losses to achieve high BCX scores as well as high predictive performances.
Furthermore, we have presented a universal variant of BCXR that improves all agreement metrics simultaneously.
By conducting experiments on eight real-world datasets, we have demonstrated that BCXR offers superior trade-offs between BCX scores and predictive performances, which underscores the effectiveness of our proposed approaches.
Overall, our study contributes to the advancement of trustworthy and responsible MLOps by providing a method to assess and enhance the consistency of explanations in model updates.

%%%%%%%%%%%%%%%%%%%%%%%%%%%%%%%%%%%%%%%%%%%%%%%%%%%%%%%%%%%%%%%%%%%%%%%%

%%% Use this command to include your bibliography file.

\bibliography{reference}

\clearpage
\onecolumn
\appendix
\section*{Appendix}
We provide the detailed proofs in \appref{proofs} and supplemental information on our numerical experiments in \appref{exp}.
Moreover, we discuss the limitation of our method in Appendix C.
%\vspace{1em}

\numberwithin{equation}{section}
\setcounter{table}{0}
\renewcommand{\thetable}{\Alph{section}.\arabic{table}}
\numberwithin{table}{section}

\setcounter{figure}{0}
\renewcommand{\thefigure}{\Alph{section}.\arabic{figure}}
\numberwithin{figure}{section}

\section{Proofs}
\applab{proofs}

\subsection{Proof of \lemref{feat}}

The inequality is trivial when $\featagree(\be_1, \be_2; k) = 1$.
Suppose $\featagree(\be_1, \be_2; k) = 1 - a / k$ with $a \in \{1 ,..., d - k\}$.
We have
\begin{align}
    \frac{1 - \featagree(\be_1, \be_2; k)}{\lfeat(\be_2; \be_1, k)} 
    &= \frac{a}{k} \frac{1}{\lfeat(\be_2; \be_1, k)} \\
    &\leq \frac{a}{k} \sup_{\be_2:\featagree(\be_1, \be_2; k) = 1- \frac{a}{k}} \frac{1}{\lfeat(\be_2; \be_1, k)} \\
    &\leq \frac{a}{k} \frac{1}{\inf_{\be_2:\featagree(\be_1, \be_2; k) = 1- \frac{a}{k}} \lfeat(\be_2; \be_1, k)} \\
    &\leq \frac{a}{k} \frac{k}{ a \varepsilon } = \varepsilon^{-1}  \eqlab{tenge}
\end{align}
where $\lfeat(\be_2; \be_1, k)$ is infimized when $|e_{2i}| = \psi_{feat}(\be_2)$ for each 
%$i \in \topfeat(\be_1;k) \setminus  \topfeat(\be_2;k)$.
%$i \in \topfeat(\be_1;k) \setminus \{ i \mid |e_{2i}| > \tau_{feat} \}$.
$i \in \{ i \in \topfeat(\be_1;k) \mid |e_{2i}| \leq \psi_{feat}(\be_2) \}$.
At this time, $\lfeat(\be_2; \be_1, k)$ is at least $a \varepsilon /k$.
Since \eqref{tenge} does not depend on $a$, $\varepsilon^{-1} \lfeat(\be_2; \be_1, k)$ always bounds $(1 - \featagree(\be_1, \be_2; k) )$ from above, concluding the proof.
\QED

\subsection{Proof of \lemref{lrank}}
The proof is almost identical with \lemref{feat}.
When $\rankagree(\be_1, \be_2) = 1$, we have $\lrank(\be_1, \be_2; k) = 0$ and  the inequality is trivial.
Suppose $\rankagree(\be_1, \be_2; k) = 1 - a / k$ with $a \in \{1,..., k\}$.
We have
\begin{align}
    \frac{1 - \rankagree(\be_1, \be_2; k)}{\lrank(\be_2; \be_1, k)}
    &\leq \frac{a}{k} \frac{1}{\inf_{\be_2:\rankagree(\be_1, \be_2; k) = 1- \frac{a}{k}} \lrank(\be_2; \be_1, k)} \leq \frac{a}{k} \frac{k}{ a \varepsilon } = \varepsilon^{-1} 
\end{align}
where the infimum of $\lrank(\be_2; \be_1, k)$ is lower bounded by $a\varepsilon / k$, which may be achieved when $|e_{2i}| = \sort(\abso(\be_2)_{i=-\varepsilon})_j$ for some $a$ pairs of $(j,i)$ in $I$, concluding the proof.
\QED

\subsection{Proof of \lemref{lsign}}
The proof is almost identical with \lemref{feat}.
The inequality is trivial when $\signagree(\be_1, \be_2; k) = 1$.
Suppose $\signagree(\be_1, \be_2; k) = 1 - a / k$ with $a \in \{1 ,..., k\}$.
\begin{align}
    \frac{1 - \signagree(\be_1, \be_2; k)}{\lsign(\be_2; \be_1, k)}
    &\leq \frac{a}{k} \frac{1}{\inf_{\be_2:\signagree(\be_1, \be_2; k) = 1- \frac{a}{k}} \lsign(\be_2; \be_1, k)} \\
\end{align}
When $ \psi_{sign}(\be_2) = 0 $, 
\begin{align}
    \lsign(\be_2; \be_1, k) = \frac{1}{k} \sum_{i\in \topfeat(\be_1;k)} \max\big(0, \varepsilon - \sign(e_{1i})e_{2i} \big)
\end{align}
Based on the fact that $\sign(e_{1i}) \neq \sign(e_{2i})$ for $a$ indices, we have $\lsign(\be_2; \be_1, k) \geq a\varepsilon / k$.
For the cases when $ \psi_{sign}(\be_2) > 0 $, $\lsign(\be_2; \be_1, k)$ is infimized when $\psi_{sign}(\be_2) = \sign(e_{1i})e_{2i}$ for some $a$ indices.
For both cases, the infimum is lower bounded by $a\varepsilon / k$ and we have
\begin{align}
    \frac{1 - \signagree(\be_1, \be_2; k)}{\lsign(\be_2; \be_1, k)} \leq \varepsilon^{-1}
\end{align}
which concludes the proof.
\QED

\subsection{Proof of \lemref{lsignedrank}}
The proof is trivial by the proofs of \lemref{lrank} and \lemref{lsign}.
\QED

\subsection{Proof of \lemref{ansdjfj}}
The inequality holds true if $\signedrankagree(\be_1, \be_2; k) = 1$.
Suppose $\signedrankagree(\be_1, \be_2; k) < 1$.
We have
\begin{align}
    \frac{1 - \signedrankagree(\be_1, \be_2; k)}{\normdisagree(\be_1, \be_2)}
    &\leq  \frac{1}{\inf_{\be_2:\signedrankagree(\be_1, \be_2; k) < 1} \normdisagree(\be_1, \be_2)} \eqlab{jfng}
\end{align} 
% The condition $\signedrankagree(\be_1, \be_2; k) < 1$ indicates that at least one of the followings holds;
% \begin{enumerate}
%     \item $\exists i \in \topfeat(\be_1;k), \rank(e_{1i}) \neq \rank(e_{2i})$.
%     \item $\exists i \in \topfeat(\be_1;k), \sign(e_{1i}) \neq \sign(e_{2i})$.
% \end{enumerate}
The infimum of $\normdisagree(\be_1, \be_2)$ under $\signedrankagree(\be_1, \be_2; k) < 1$ is achievable when exactly one of the following two proposition holds;
\begin{enumerate}
    \item for $(i,j) = \argmin_{(i \neq j \in \mathrm{TopFeatures}(e_1, \max(k+1, d)))} \bigl||e_{1i}| - |e_{1j}|\bigr| $,  
    %$e_{2i} = e_{2j} = \alpha e_{1i}+ (1- \alpha)e_{1j}$ for some $ \alpha \in [0,1]$
    $|e_{2i}| = |e_{2j}| = (|e_{1i}| + |e_{1j}|)/2$
    and for any other $t \in [d] \setminus \{i,j\}, e_{2t} = e_{1t}$.
    \item $k=d$ and for $i = \argmin_i |e_{1i}|$, $e_{2i} = 0$ and for any other $t \in [d] \setminus \{i\}, e_{2t} = e_{1t}$.
\end{enumerate}
When the first proposition holds, $\normdisagree$ is lower bounded as 
\begin{align}
    \normdisagree(\be_1, \be_2)
    &= \sqrt{
        \left(|e_{1i}| - \frac{|e_{1i}| + |e_{1j}|}{2}\right)^2 + \left(|e_{1j}| - \frac{|e_{1i}| + |e_{1j}|}{2}\right)^2
    }= \sqrt{
        \frac{\left(|e_{1i}| - |e_{1j}|\right)^2 }{2}
    } \geq \sqrt{  \frac{\left(\sqrt{2}\delta\right)^2}{2}} = \delta,
\end{align}
and for the second proposition, $\normdisagree$ is bounded as 
\begin{align}
    &\normdisagree(\be_1, \be_2) = \sqrt{
        (|e_{1i}| - 0)^2
    } =|e_{1i}| \geq \delta
\end{align}
Hence, under the condition that $\signedrankagree(\be_1, \be_2; k) < 1$, $\normdisagree(\be_1, \be_2)$ is no less than $\delta$.
By \eqref{jfng}, we have
\begin{align}
    \frac{1 - \signedrankagree(\be_1, \be_2; k)}{\normdisagree(\be_1, \be_2)} \leq \delta^{-1},
\end{align}
which conclude the proof.
\QED

\section{Sensitivity against $\lambda$}
\begin{figure*}[t]
    \begin{center}
    % \centering
    \subfigure[space-ga]{\label{fig:ab}\includegraphics[width=\textwidth]{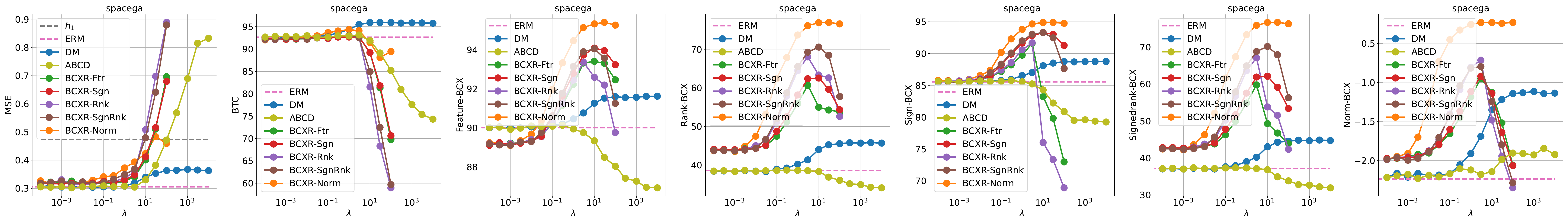}}
    % \subfigure[abalone]{\includegraphics[width=\textwidth]{fig/abalone.pdf}}
    \subfigure[cadata]{\includegraphics[width=\textwidth]{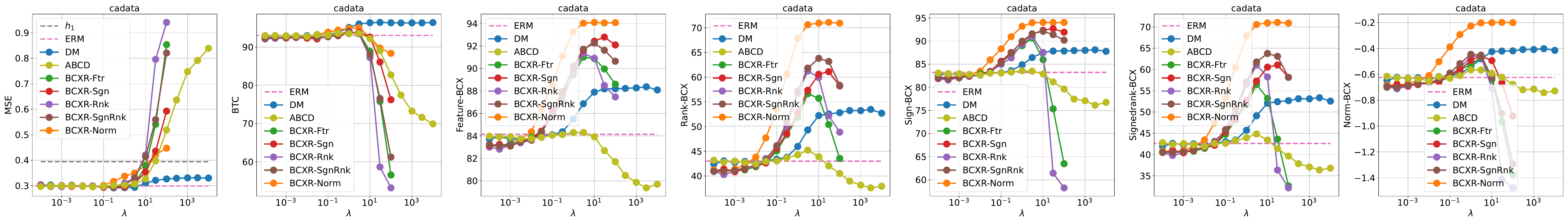}}
    \subfigure[cpusmall]{\includegraphics[width=\textwidth]{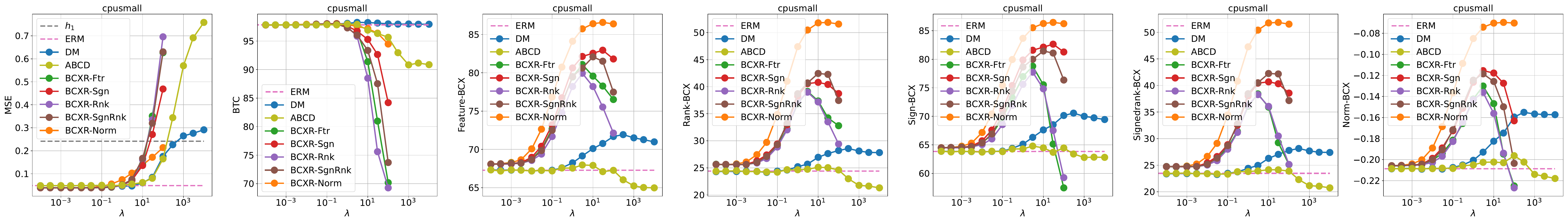}}
    \subfigure[YearPredictionMSD]{\includegraphics[width=\textwidth]{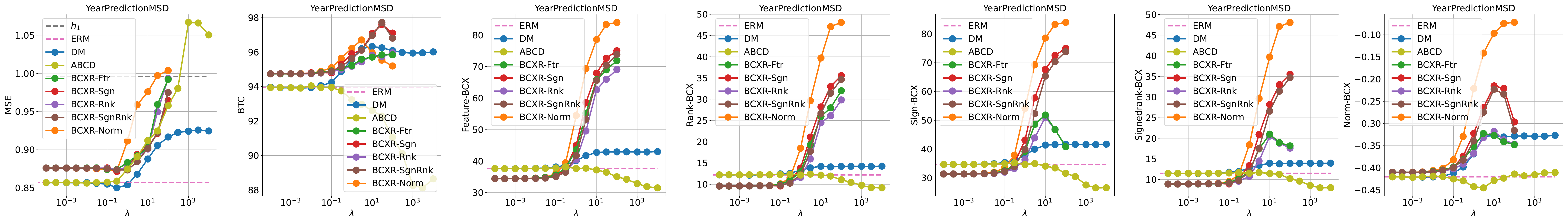}}
    \caption{Sensitivity plot of $\lambda$ for regression data sets.}
    \label{fig:fig4}
    \end{center}
\end{figure*}

\begin{figure*}[t]
    \begin{center}
    % \centering
    % \subfigure[svmguide1]{\includegraphics[width=0.95\textwidth]{fig/svmguide1.pdf}}
    \subfigure[cod-rna]{\includegraphics[width=0.95\textwidth]{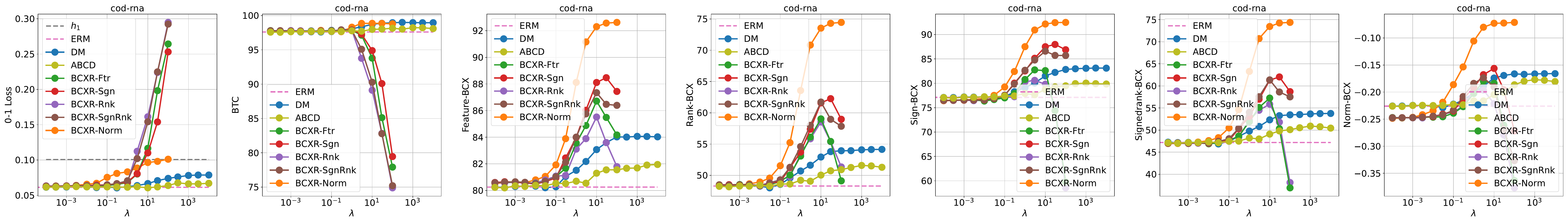}}
    % \subfigure[ijcnn1]{\includegraphics[width=0.95\textwidth]{fig/ijcnn1.pdf}}
    \subfigure[phishing]{\includegraphics[width=0.95\textwidth]{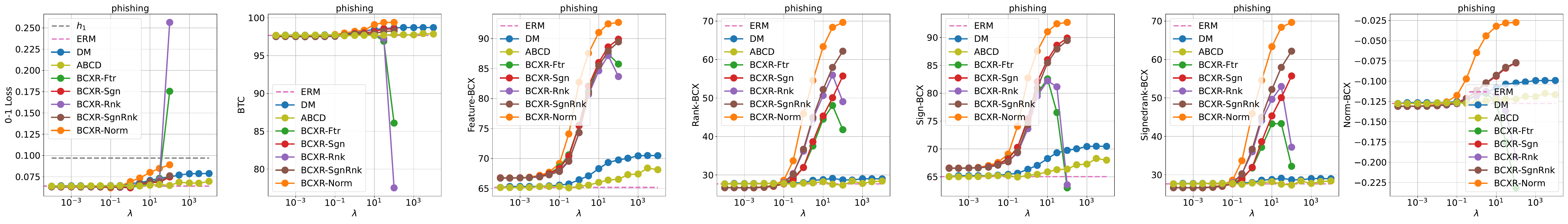}}
    % \subfigure[mushrooms]{\includegraphics[width=0.95\textwidth]{fig/mushrooms.pdf}}
    \subfigure[a9a]{\includegraphics[width=0.95\textwidth]{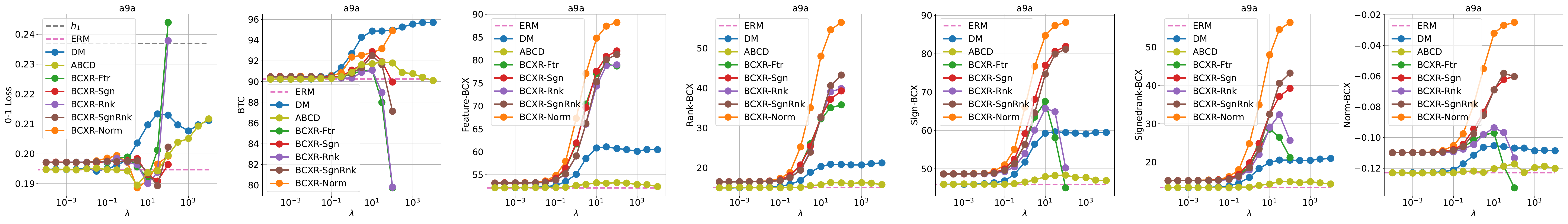}}
    \subfigure[w8a]{\includegraphics[width=0.95\textwidth]{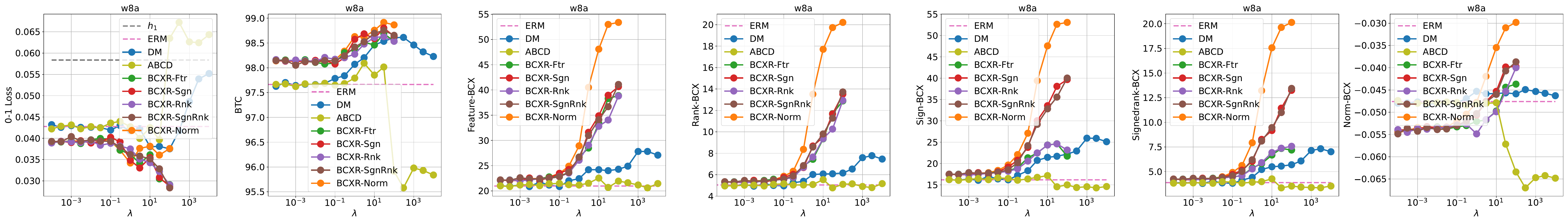}}
    \caption{Sensitivity plot of $\lambda$ for classification data sets.}
    \label{fig:fig5}
    \end{center}
\end{figure*}

\applab{exp}
While we have provided BCX-against-loss plots in our main paper, we have also included additional plots that illustrate the results against the hyperparameters $\lambda$ for each data set and each metric.
The plots are presented in \figref{fig4} for regression tasks and \figref{fig5} for  classification tasks.
The results clearly indicate that the agreement-based BCXR methods are highly sensitive to changes in the value of $\lambda$.
For example, when $\lambda$ is set to $10^2$ for the space-ga data set, the mean squared error (MSE) of the BCXR methods often exceeds the MSE of the old models.
A lower MSE than the old model is crucial for successful model updates, and therefore, setting $\lambda$ to a large value can adversely affect the training of a new model.
However, based on these findings, we can conclude that setting $\lambda$ to a value between 1 and 10 would generally yield good results for most data sets.
Therefore, when applying our BCXR method in practical tasks, it is recommended to tune $\lambda$ within this range to ensure the training of a suitable model in MLOps.

% For data sets with smaller number of features such as space-ga ($d=6$), cadata ($d=9$) and cod-rna ($d=8$)
% \begin{figure*}[h]
% \centering
% \includegraphics[width=\textwidth]{fig/sample.pdf}
% \caption{Results for cod-rna (classification) data set.}
% \label{fig:eurai}
% \end{figure*}

\section{Limitation}
In this section, we discuss the possible limitation of BCX and BCXR.

% \block{Usefulness of BCX and BCXR.}
% For example, in a pathology model that predicts disease names based on patient symptoms, the explanation of the model may be heavily used to advise doctors and explain the predictions to patients.
% Suppose the explanations are inconsistent among pre- and post-update models.
% Then, even if the predictions of the post-update model are correct, it will lose the trust of doctors and patients, and this will be a real issue in practical MLOps.
% Compatibility of explanation is important because these issues will surely arise in areas where responsible AI is required, such as healthcare, finance, and security.

\block{Intractable computational cost.}
BCX and BCXR may suffer from intractable computational costs due to the high complexity of the explanation methods they employ.
For example, in our experiments, we utilize SHAP.
However, as the number of features $d$ increases, SHAP becomes increasingly computationally intensive.
Therefore, when dealing with a very large number of features (e.g., in image and text classification), it is preferable to approximate the SHAP calculation or to use more lightweight explanation methods, such as gradient-based methods~\cite{gradcam2020, simonyan2014deep,integratedgrad2017sundararajan}.
Although these alternatives may sacrifice some of the validity of the explanation, they provide a more computationally feasible solution.

\block{Difficulty of requirement design.}
Although BCX quantitatively assesses the consistency of explanations, determining the practical requirements for BCX in real-world applications can be challenging.
Both BCX and BCXR are mathematically defined metrics, which means that outliers, abnormal explanations, and distribution shifts do not affect their computation, as is the case with any BTC-related scores.
However, we observe that these scores might be practically meaningless in certain contexts.
For example, if a pre-update model is trained before a severe distribution shift, aligning a post-update model with the pre-update model may not be reasonable or useful.
The meaningfulness of BCX and other BTC-related scores depends on the compatibility requirements for the ML system.
Unfortunately, these requirements cannot be uniquely determined from a theoretical perspective alone.
Therefore, data scientists and customers need to collaboratively discuss the detailed requirements to determine the necessary level of compatibility for different situations.
Based on these requirements, it may be necessary to remove outliers and abnormal explanations from the computation of BCX.
Although determining these requirements may limit the practical usefulness of BCX and BCXR, this challenge is common in the broader fields of explainability, fairness, and privacy in machine learning.

\end{document}